\pdfoutput=1

\documentclass[11pt]{article}

\usepackage[preprint]{acl}

\usepackage{times}
\usepackage{algorithm}
\usepackage{algorithmic}
\usepackage{arydshln}
\usepackage{amsmath}
\usepackage{mathtools}
\usepackage{amssymb}
\usepackage{amsfonts}
\usepackage{bm}
\usepackage{bbm}
\usepackage{booktabs}
\usepackage{multirow}
\usepackage{graphicx}
\usepackage{tabularx}
\usepackage{xcolor}
\usepackage{url}
\usepackage{siunitx}
\usepackage{booktabs}
\usepackage{makecell}
\usepackage{etoolbox}
\usepackage{caption}
\usepackage{subcaption}
\usepackage{comment}
\usepackage{latexsym}

\usepackage{xspace}
\usepackage{color, colortbl}

\definecolor{LightCyan}{rgb}{0.88,1,1}

\usepackage[T1]{fontenc}

\usepackage[utf8]{inputenc}

\usepackage{microtype}

\usepackage{inconsolata}

\usepackage{graphicx}

\title{Efficient Nearest Neighbor based Uncertainty Estimation for Natural Language Processing Tasks}

\author{Wataru Hashimoto, \
  Hidetaka Kamigaito, \
  Taro Watanabe \\
  Nara Institute of Science and Technology \\
  \texttt{\{hashimoto.wataru.hq3, kamigaito.h, taro\}@is.naist.jp}}

\begin{document}
\maketitle
\begin{abstract}
Trustworthiness in model predictions is crucial for safety-critical applications in the real world.
However, deep neural networks often suffer from the issues of uncertainty estimation, such as miscalibration.
In this study, we propose $k$-Nearest Neighbor Uncertainty Estimation ($k$NN-UE), which is a new uncertainty estimation method that uses not only the distances from the neighbors, but also the ratio of labels in the neighbors.
Experiments on sentiment analysis, natural language inference, and named entity recognition show that our proposed method outperforms the baselines and recent density-based methods in several calibration and uncertainty metrics.
Moreover, our analyses indicate that approximate nearest neighbor search techniques reduce the inference overhead without significantly degrading the uncertainty estimation performance when they are appropriately combined.
\end{abstract}

\section{Introduction}
In order to deploy Deep Neural Networks (DNNs) including Pre-trained Language Models (PLMs) in safety-critical areas, uncertainty estimation (UE) is important.
Improving the predictive uncertainty will calibrate the prediction~\cite{pmlr-v70-guo17a},\footnote{"Calibration" means the confidence of the prediction aligns with its accuracy.} or enhance the selective prediction performance which reduces incorrect predictions by providing the option to abstain from the model prediction~\cite{galil2023what}.
On the other hand, DNNs often fail to quantify the predictive uncertainty, for example, causing miscalibrated prediction~\cite{pmlr-v70-guo17a}.
Such UE performance problems can be mitigated by the PLMs, such as BERT~\cite{devlin-etal-2019-bert} or DeBERTa~\cite{he2021deberta}, that are self-trained on vast amounts of data~\cite{ulmer-etal-2022-exploring}; nevertheless, there remains considerable room for improvement~\cite{desai-durrett-2020-calibration}. \par

\begin{figure}[t!]
    \includegraphics[width=0.48\textwidth]{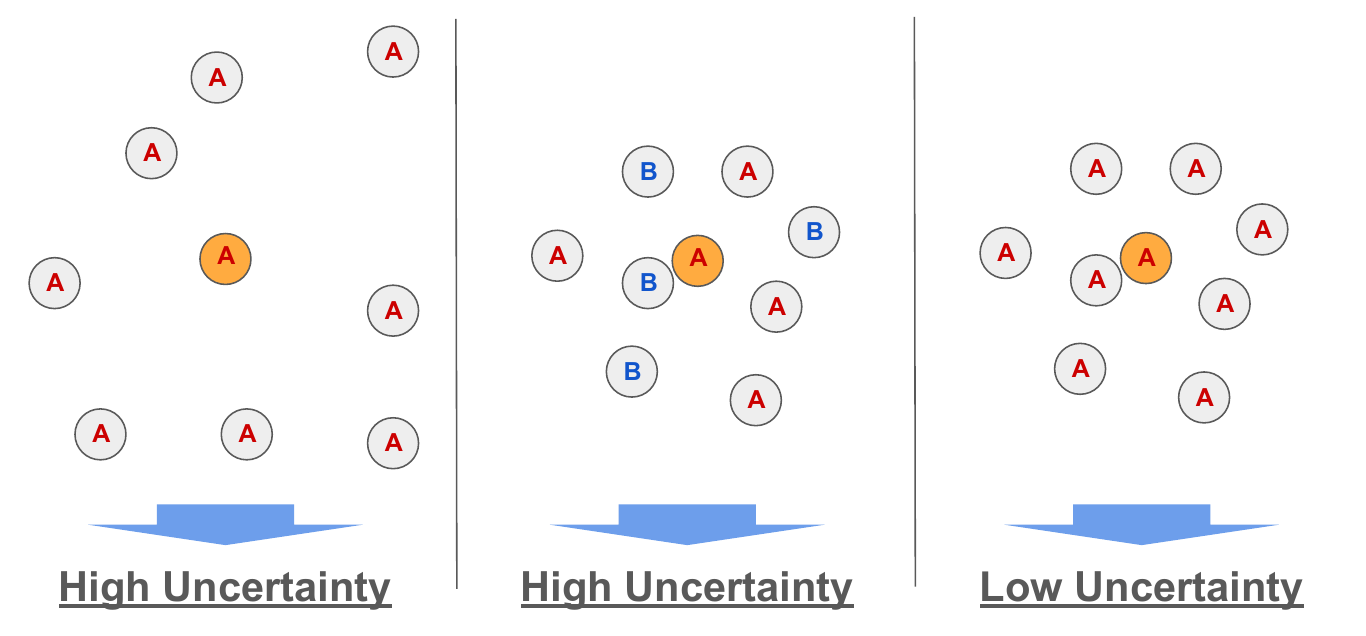}
    \caption{Illustrations of $k$NN-UE behavior. The orange circle indicates predicted data instances and other circles indicate training data instances. $k$NN-UE gives high uncertainty when the predicted query representation is far from examples obtained from the $k$NN search (left) and the predicted label is different from the labels of neighbors (center). $k$NN-UE outputs low uncertainty only when the query representation is close to neighbors and the labels of neighbors contain many of the model's predicted label (right).}
    \label{fig:overview_of_knn_ue}
\end{figure}

To address the challenge of UE, multiple stochastic inferences such as MC Dropout~\cite{pmlr-v48-gal16} and Deep Ensembles~\cite{deep-ensemble-2017} are generally effective.
On the other hand, these methods require multiple stochastic inferences for a single data instance, which leads to high computational cost, and makes them impractical for real world application.
To balance reasonable predictive uncertainty with computational efficiency, Temperature Scaling~\cite{pmlr-v70-guo17a} , which scales logits by a temperature parameter, is commonly employed.
Furthermore, density-based methods, such as Density Softmax~\cite{pmlr-v235-bui24a} and Density Aware Calibration (DAC)~\cite{Tomani-dac-2023}, have demonstrated promising UE performance and inference costs by adjusting model outputs based on estimated density.

However, both Density Softmax and DAC only use the density of the training data.
Relying on density alone can sometimes lead to overconfident predictions, even when such confidence is unwarranted.
For instance, the neighbor of the input may contain many examples with labels that differ from the predicted label.
In this situation, the prediction should obviously not be trusted.
Therefore, we hypothesized that considering both the density and the label information of the neighbors will improve UE performance.


In this study, we propose $k$-Nearest Neighbor Uncertainty Estimation ($k$NN-UE), a new density-based UE method that reflects nearest neighbor labels.
As illustrated in Figure~\ref{fig:overview_of_knn_ue}, $k$NN-UE is designed to achieve the highest prediction confidence when the input and the nearest neighbors are both close in distance and share the same label as the predicted label.
Our method weights logits according to the score from the distance between the input example and its neighbors in the datastore created by the training data and the ratio of the model's predicted label matched with the labels in the neighbors.
In addition, our method requires only a single forward inference of the model with almost no additional computational cost.
The contributions of this research are as follows.

First, our experiments show that $k$NN-UE improves the UE performance of existing baselines in sentiment analysis, natural language inference, and named entity recognition in both in-domain and out-of-domain settings by combining neighbor label information and distances from neighbors.
On the other hand, we also find that naive $k$NN-UE makes less efficient for token-level tasks such as \textit{sequence-labeling} based named entity recognition due to the execution of $k$NN to each token.

Second, to mitigate the above latency problem in $k$NN-UE, we show that approximate $k$NN search or dimension reduction in $k$NN-UE improves the inference speed without degrading UE performance much more, while combining them leads to degrading the UE performance.

Our code is available at \url{https://github.com/wataruhashimoto52/knn_ue}.

\section{Related Work}

\paragraph{Uncertainty Estimation for Natural Language Processing Tasks} 
Studies about UE for NLP tasks are limited when compared with those for image datasets.~\citet{kotelevskii2022nonparametric} has shown excellent performance in classification with rejection tasks and out-of-distribution detection tasks using uncertainty scores using density estimation results. ~\citet{vazhentsev-etal-2022-uncertainty} performed misclassification detection using Determinantal point processes~\cite{dpp_2012}, spectral normalization, Mahalanobis distance and loss regularization in text classification and NER.
However, these are still focusing only on the feature representation or the density, not the labels of the neighbors. 
\citet{he-etal-2024-uncertainty} proposed a framework that considers uncertainty between tokens in NER. However, the target task is limited to NER, and it is not for confidence calibration.
\citet{hashimoto-etal-2024-data} shows that simple data augmentation methods in NER can improve UE performance without additional inference costs, but its effectiveness is limited in the in-domain.
Our $k$NN-UE improves the UE performance in the in-domain and the out-of-domain classification and NER tasks using not only $k$NN density but also neighbor labels.

\paragraph{$k$-Nearest Neighbor Language Models / Machine Translation}
$k$-Nearest Neighbor Language Model ($k$NN-LM)~\cite{Khandelwal2020Generalization} has been proposed, which performs linear interpolation of $k$NN probability based on distance from neighbors and base model probability, in the language modeling task.
$k$-Nearest Neighbor Machine Translation ($k$NN-MT) applied the $k$NN-LM framework to machine translation~\cite{khandelwal2021nearest}.
$k$NN-LM and $k$NN-MT have been successful because they enhance predictive performance through the memorization and use of rich token representations of pre-trained language models and mitigate problems such as a sparsity comes from low-frequency tokens~\cite{zhu-etal-2023-ink}.
The main issue on $k$NN-LM and $k$NN-MT is the inference overhead, and there are several studies to solve this problem. 
\citet{he-etal-2021-efficient} employs datastore compression, adaptive retrieval, and dimension reduction to reduce computational overhead with retaining perplexity.
\citet{deguchi-etal-2023-subset} dramatically improves decoding speed by dynamically narrowing down the search area based on the source sentence.
We investigate that whether UE performance in $k$NN-UE can keep or not with reducing inference time by introducing some of the speed-up techniques established in kNN-LM/MT.

\section{Preliminary}
\label{sec:preliminary}

\subsection{Definitions}
In multiclass classification, we assume a dataset \( \mathcal{D} = \{(\bm{x}_n, y_n)\}_{n=1}^N \) consisting of \( N \) examples, where \( y_n \in \{1, 2, \ldots, J\} \) denotes its corresponding class label among \( J \) possible classes.\footnote{In the case of sequence labeling, we can interpret the number of data $N$ as the product of the raw number of data instances and the sequence length.} We use the trained neural network feature extractor $f$ and the classifier $g$ for classification, where $f(\bm{x}) \in \mathbb{R}^D$. $g$ gives us the logits $\bm{z} = g(f(\bm{x}))$ and we obtain the confidence $\bm{p} = \mathrm{softmax} (\bm{z})$.

\subsection{Density Softmax}
Density Softmax~\cite{pmlr-v235-bui24a} obtains confidence by weighting logits with normalized log-likelihood from a trained density estimator.
$\bm{\beta}$ are the parameters of the density estimator; $p(f(\bm{x}); \bm{\beta})$ is the normalized log-likelihood from the density estimator, then the corrected confidence is written as
\begin{equation}
    p(y_{i}|\bm{x}) = \frac{{\rm{exp}} \left(p(f(\bm{x}); \bm{\beta}) \cdot z_{i}\right)}{\sum_{j=1}^J {\rm{exp}} \left(p(f(\bm{x}); \bm{\beta}) \cdot z_{j} \right)}. \\
\end{equation}

In Density Softmax, the closer the normalized log-likelihood to zero, the closer the prediction to Uniform distribution.
Density Softmax achieves reasonable latency and competitive UE performance with state-of-the-art methods at the cost of demanding the density estimator training and multiple base model training.\footnote{Details for the density estimator in this study are in Appendix~\ref{sec:appendix_training_density_estimator}.}

\subsection{Density Aware Calibration (DAC)}
DAC is a confidence calibration method using multiple feature representations, which is similar to the $k$NN-based out-of-distribution detection~\cite{pmlr-v162-sun22d}.
DAC~\cite{Tomani-dac-2023} scales the logits by using sample-dependent temperature $\Phi(\bm{x}, \bm{w})$
\begin{align}
    p(y_{i}|\bm{x}) &= \frac{ {\rm{exp}} \left( z_{i} / {\Phi(\bm{x}, \bm{w})} \right)}{\sum_{j=1}^J {\rm{exp}} \left( z_{j} / {\Phi(\bm{x}, \bm{w})} \right)} \\
    \text{where} \notag \\ 
    \Phi(\bm{x}, \bm{w}) &= \sum_{l=1}^L w_{l}s_{l} + w_{0}.
\end{align}
$\bm{w} \in {w_{1}...w_{L}}$ are the weights for every layer of the base model, $s_{l}$ is the averaged distance from $k$NN search on $l$-th layer, and $w_{0}$ is the bias term. $w_{0}...w_{L}$ are optimized using the L-BFGS-B method~\cite{Liu1989OnTL} based on the loss in the validation set.
In the original DAC paper, the UE performance tends to improve with the increase in the number of layer representation~\cite{Tomani-dac-2023}.
Therefore, we use all the hidden representations in each layer of the base PLMs.

DAC is a non-parametric method that makes not assumptions about the training data distribution unlike Density Softmax~\cite{pmlr-v235-bui24a}, which relies on some density estimators.
On the other hand, the recent $k$NN-based DAC still relies only on the distances to the neighbors.
These methods do not take into account the label information of the input neighbors, which limits the improvement of the UE performance.


\section{Proposed Method: \texorpdfstring{$k$}{k}-Nearest Neighbor Uncertainty Estimation (\texorpdfstring{$k$}{k}NN-UE)}
\label{sec:knn_ue_explanation}

\begin{figure*}[t!]
    \centering
    \includegraphics[width=14.2cm]{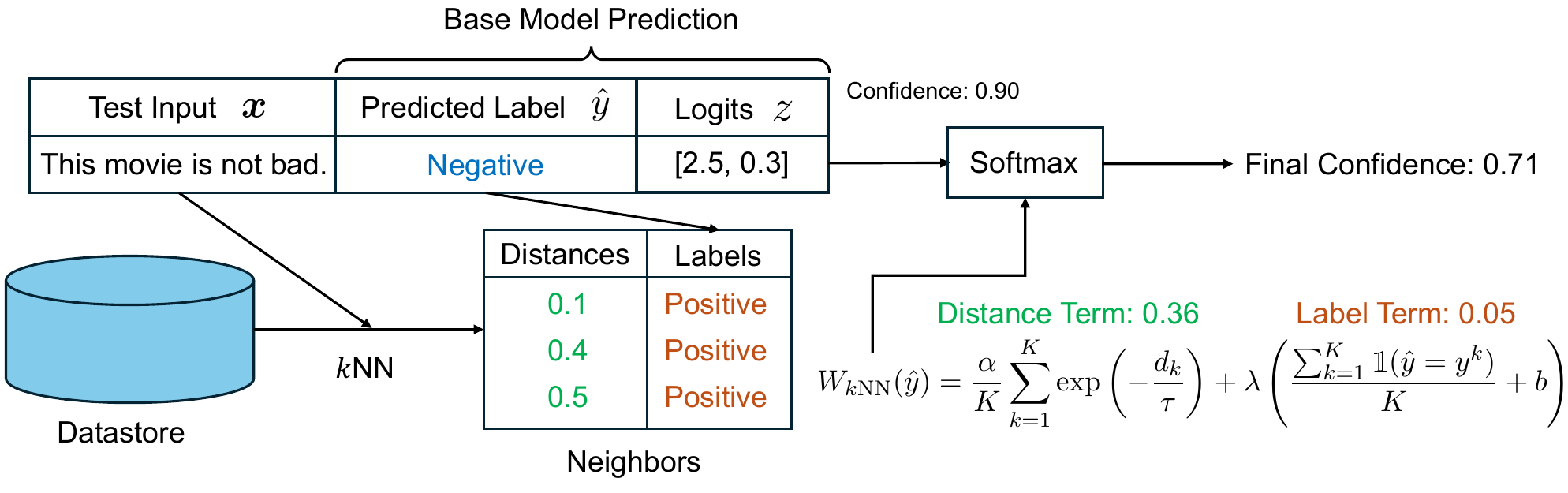}
    \caption{A diagram of $k$NN-UE when $K = 3$ and the estimated hyperparameters are $\alpha = 0.5$, $\tau = 1.0$, $\lambda = 0.5$ and $b = 0.1$. A datastore is constructed with the representations of the training data as keys and their labels as values. The distances of the nearest examples from the test representation, and the neighbor labels are aggregated into $W_{k {\rm{NN}}} (\hat{y})$. Finally we obtain calibrated confidence by correcting the raw logits with $W_{k {\rm{NN}}} (\hat{y})$ as in Eq. ~\ref{eq:knn_ue_main}.}
    \label{fig:diagram_knn_ue}
\end{figure*}

The main idea of our proposed method, $k$NN-UE, stems from the notion that the density-based UE methods can be further improved by using label information about the training data instances that make up the density.

In order to take into account the variance of neighbor labels, our $k$NN-UE explicitly includes the label agreement information of the predicted instance and its neighbor examples when calculating the confidence.
More specifically, we regard the prediction as more reliable only when the prediction is in a region where training data is dense and the predicted label and the labels of the data instances that make up the dense region are mostly the same, as illustrated in the right part of Figure~\ref{fig:overview_of_knn_ue}.
Otherwise, for example, if there is a lot of discrepancy in the neighbor labels and the predicted label, we treat the prediction as unreliable, indicated in the middle of Figure~\ref{fig:overview_of_knn_ue}.


In our $k$NN-UE, we introduce two terms: one related to the density of the training data and one related to the degree of agreement of the predicted data and neighbor labels. Confidence of $i$-th label obtained by $k$NN-UE is following the formula:


\begin{align}
\label{eq:knn_ue_main}
    p(y_{i}|\bm{x}) &= \frac{ {\rm{exp}} ( W_{k {\rm{NN}}} (\hat{y}) \cdot z_{i})}{\sum_{j=1}^J {\rm{exp}} ({ W_{k {\rm{NN}}} (\hat{y}) \cdot z_{j}})} \\
    \text{where} \notag \\
    W_{k {\rm{NN}}} (\hat{y}) &= \underbrace{\frac{\alpha}{K} \sum_{k=1}^{K}{\rm{exp}} \left( -\frac{d_k}{\tau}\right)}_{\text{distance term}} \notag \\
    &+ \underbrace{\lambda \left(\frac{S(\hat{y})}{K} + b \right)}_{\text{label term}}.
    \label{eq:knn_ue_weight}
\end{align}
$K$ is the number of neighbors from $k$NN search, $S(\hat{y}) = \sum_{k=1}^{K} \mathbbm{1} (\hat{y} = y^{k})$ is the count when the predicted label $\hat{y}$ and the label of the $k$-th neighbor $y^{k}$ is same, $d_k$ is the distance between the $k$-th $f(\bm{x})$ representation obtained by $k$NN search and the representations of training data.\footnote{Note that $k$NN-UE is also "accuracy-preserving" same as DAC because $W_{k {\rm{NN}}} (\hat{y})$ is a scalar, not a class-wise score.}
The parameters $\alpha$, $\tau$, $\lambda$ $\in \mathbb{R}_{+}$ and $b$ $\in \mathbb{R}$ are optimized using the L-BFGS-B method based on the loss in the validation set.

When the distance and label terms are smaller and $W_{k {\rm{NN}}} (\hat{y})$ is closer to zero, the closer the prediction is to Uniform distribution, which allows us to better estimate the confidence of the prediction.
In this study, we also conduct experiments without the label term in Equation~\ref{eq:knn_ue_weight}, to emphasize the importance of $k$NN neighbor labels in UE. We summarize a diagram of $k$NN-UE in Figure~\ref{fig:diagram_knn_ue}.





\section{Experimental Settings}
\label{sec:experimental_settings}

\subsection{Tasks and Datasets}
\label{sec:tasks_datasets}
We measure the UE performance on Sentiment Analysis (SA), Natural Language Inference (NLI), and Named Entity Recognition (NER) in In-domain (ID) and Out-of-Domain (OOD) settings.\footnote{The datasets in SA and NLI were set up with reference to ~\citet{xiao-etal-2022-uncertainty}.} Dataset statistics are described in Appendix~\ref{sec:appendix_dataset_statistics}.

\paragraph{Sentiment Analysis (SA)} is a task to classify whether the text sentiment is positive or negative. The IMDb movie review dataset~\cite{maas-etal-2011-learning} is treated as ID, and the Yelp restaurant review dataset~\cite{yelp_NIPS2015} is treated as OOD.
\paragraph{Natural Language Inference (NLI)} classifies the relationship between a hypothesis sentence and a premise sentence. We treat the Multi-Genre Natural Language Inference (MNLI) dataset~\cite{williams-etal-2018-broad} as ID and the Stanford Natural Language Inference (SNLI) dataset~\cite{bowman-etal-2015-large} as OOD.
\paragraph{Named Entity Recognition (NER)} extracts the named entities, such as a person, organization, or location. The NER task was carried out in the framework of \textit{sequence labeling}. We regard the OntoNotes 5.0 dataset~\cite{pradhan-etal-2013-towards} broadcast news ($\mathtt{bn}$) domain as ID, and newswire ($\mathtt{nw}$) and telephone conversation ($\mathtt{tc}$) domains as OOD.

\subsection{Existing Methods}
\label{sec:existing_methods}
We employ the simple baselines: Softmax Response (SR)~\cite{1995SR}, Temperature Scaling (TS)~\cite{pmlr-v70-guo17a}, Label Smoothing~\cite{Miller1996AGO, pereyra2017regularizing} and MC Dropout~\cite{pmlr-v48-gal16}. In addition, we use the recent strong baselines for UE: Spectral-Normalized Gaussian Process (SNGP)~\cite{NEURIPS2020_SNGP}, Posterior Networks (PN)~\cite{posterior_network_NEURIPS2020}, Mahalanobis Distance with Spectral-Normalized Network (MDSN)~\cite{vazhentsev-etal-2022-uncertainty}, E-NER~\cite{zhang-etal-2023-ener}, Density Softmax~\cite{pmlr-v235-bui24a}, and DAC~\cite{Tomani-dac-2023}. Details on baselines can be found in Appendix~\ref{sec:appendix_details_baselines}. We have also experimented with a variant of $k$NN-UE without the label term in Eq. \ref{eq:knn_ue_weight}, denoted by ``w/o label'' to emphasize the impact of the neighbor labels.

\subsection{Training Settings}
\label{sec:train_settings}

In all experiments, we train and evaluate the models on a single NVIDIA A100 GPU with 40GB of memory. We used $\mathrm{DeBERTaV3_{BASE}}$\footnote{ \url{https://huggingface.co/microsoft/deberta-v3-base}} and $\mathrm{mDeBERTaV3_{BASE}}$\footnote{\url{https://huggingface.co/microsoft/mdeberta-v3-base}} \cite{he2023debertav}, as the Transformer encoder from $\mathtt{transformers}$ \cite{wolf-etal-2020-transformers} pre-trained model checkpoints.
We use the cross-entropy loss in all experiments, including the optimization of hyperparameters in $k$NN-UE.
Batch size is 32, and the initial learning rate was set to 1e-5.
The gradient clipping is applied with the maximum norm of 1.
All experiments are run five times, and we report the mean and standard deviation of the scores.

\paragraph{Datastore Construction}
It is necessary to maintain the representation of the data for training a density estimator in Density Softmax and $k$NN search in DAC and $k$NN-UE. We use the final layer representations corresponding to CLS tokens in SA and NLI. In NER, we stored the hidden representation of the final layer as a token representation corresponding to the beginning of the word.

\paragraph{$k$-Nearest Neighbor Search}
We use $\mathtt{faiss}$~\cite{douze2024faiss} as the GPU-accelerated $k$NN search toolkit. Unless otherwise specified, we fix the number of neighbors $K = \mathrm{32}$ in $k$NN search,\footnote{In Section~\ref{sec:impact_of_top_k}, we conducted experiments to examine the behavior when varying $K$ over the set $\{8, 16, 32, 64, 128\}$, with $K = 32$ representing the median.} and use $\mathtt{faiss.IndexFlatL2}$ which is an index for exact search in L2 norm, as the default in $k$NN-UE.

\begin{table*}[t]
\centering
\scalebox{0.68}{
\begin{tabular}{l|cccc|cccc}
    \hline
    \multicolumn{1}{l|}{Methods} & \multicolumn{4}{c|}{IMDb (In-domain)} & \multicolumn{4}{c}{Yelp (Out-of-domain)} \\ \cline{2-9}
     & ECE (↓) & MCE (↓) & AUROC (↑) & E-AURC (↓) & ECE (↓) & MCE (↓) & AUROC (↑) & E-AURC (↓) \\ \hline
    SR & 4.42$\pm$0.41 & 24.06$\pm$3.52 & 98.35$\pm$0.10 & 10.60$\pm$2.81 & 4.69$\pm$1.20 & 21.02$\pm$6.74 & 98.15$\pm$0.39 & 11.84$\pm$3.15 \\
    TS & 4.10$\pm$0.31 & 20.43$\pm$5.01 & 98.45$\pm$0.21 & 11.36$\pm$2.82 & 5.10$\pm$1.19 & 19.70$\pm$1.35 & 98.20$\pm$0.46 & 12.91$\pm$4.12 \\
    LS & 1.88$\pm$0.41 & 21.50$\pm$4.53 & 98.36$\pm$0.45 & 14.52$\pm$7.24 & 2.53$\pm$0.43 & 16.47$\pm$3.51 & \textbf{98.30$\pm$0.45} & 12.90$\pm$6.09 \\
    MC Dropout & 4.28$\pm$0.27 & 23.74$\pm$3.52 & 98.57$\pm$0.12 & 9.17$\pm$1.74 & 4.33$\pm$0.54 & 20.17$\pm$2.79 & 98.28$\pm$0.25 & 10.01$\pm$2.01 \\
    SNGP & 4.18$\pm$0.30 & 22.69$\pm$4.83 & 98.53$\pm$0.15 & 9.95$\pm$1.17 & 4.89$\pm$0.59 & 21.28$\pm$4.68 & 98.10$\pm$0.27 & 11.42$\pm$2.14 \\
    PN & 4.28$\pm$0.43 & 24.43$\pm$0.20 & 98.06$\pm$0.27 & 10.99$\pm$5.63 & 4.69$\pm$0.35 & 24.41$\pm$0.32 & 97.56$\pm$0.25 & 15.82$\pm$3.94 \\
    MDSN & 4.45$\pm$0.43 & 23.97$\pm$5.05 & 98.48$\pm$0.08 & 10.25$\pm$0.86 & 5.32$\pm$0.92 & 21.33$\pm$2.91 & 98.00$\pm$0.20 & 11.12$\pm$3.53 \\
    Density Softmax & 4.23$\pm$0.36 & 27.10$\pm$6.92 & 98.34$\pm$0.08 & 11.39$\pm$2.48 & 4.99$\pm$0.48 & 21.98$\pm$3.68 & 98.09$\pm$0.24 & 13.05$\pm$2.72 \\
    DAC & 1.51$\pm$0.33 & 14.17$\pm$2.73 & 98.36$\pm$0.37 & 12.72$\pm$6.15 & 2.35$\pm$0.12 & 6.44$\pm$2.23 & 97.86$\pm$0.60 & 14.26$\pm$5.90 \\ \hline
    $k$NN-UE (w/o label) & 1.33$\pm$0.36 & 13.13$\pm$3.24 & \textbf{98.65$\pm$0.13} & 9.36$\pm$0.36 & 2.23$\pm$0.29 & 6.33$\pm$2.76 & 98.27$\pm$0.11 & 10.97$\pm$0.91 \\
    $k$NN-UE  & \textbf{0.95$\pm$0.12}\textsuperscript{†} & \textbf{9.02$\pm$1.39}\textsuperscript{†} & 98.64$\pm$0.12 & \textbf{7.97$\pm$0.61}\textsuperscript{†} & \textbf{1.45$\pm$0.15}\textsuperscript{†} & \textbf{4.17$\pm$1.52} & 98.23$\pm$0.39 & \textbf{9.92$\pm$0.61} \\ \hline
\end{tabular}
}
\caption{ECE, MCE, AUROC, and E-AURC results about SA task on IMDb (In-domain) and Yelp (Out-of-domain) for $\mathrm{mDeBERTaV3_{BASE}}$ model. Bolds indicate the best result. {\textdagger} indicates significantly improved than existing methods (p < 0.05) by using t-test.}
\label{tab:sentiment_analysis_metrics}
\end{table*}

\begin{table*}[t]
\centering
\scalebox{0.68}{
\begin{tabular}{l|cccc|cccc}
\hline
\multicolumn{1}{l|}{Methods} & \multicolumn{4}{c|}{MNLI (In-domain)} & \multicolumn{4}{c}{SNLI (Out-of-domain)} \\ \cline{2-9}
     & ECE (↓) & MCE (↓) & AUROC (↑) & E-AURC (↓) & ECE (↓) & MCE (↓) & AUROC (↑) & E-AURC (↓) \\ \hline
SR & 8.36$\pm$0.61 & 37.61$\pm$7.53 & 97.03$\pm$0.12 & 31.29$\pm$2.23 & 9.77$\pm$0.55 & 36.61$\pm$14.05 & 96.07$\pm$0.17 & 37.62$\pm$0.67 \\
TS & 2.73$\pm$1.86 & 15.81$\pm$11.05& 97.06$\pm$0.02 & 31.24$\pm$1.86 & 3.92$\pm$1.79 & 18.13$\pm$10.69 & 96.08$\pm$0.13 & 38.40$\pm$2.06 \\
LS & 2.89$\pm$0.14 & 28.64$\pm$7.90 & 96.56$\pm$0.55 & 37.98$\pm$12.64 & 3.97$\pm$0.45 & 23.18$\pm$6.17 & 95.61$\pm$0.40 & 44.18$\pm$9.18 \\
MC Dropout & 8.13$\pm$0.65 & 30.17$\pm$6.83 & 96.97$\pm$0.06 & 32.31$\pm$2.25 & 9.62$\pm$0.53 & 28.90$\pm$5.03 & 96.10$\pm$0.11 & 37.19$\pm$2.99 \\
SNGP & 10.45$\pm$0.56 & 35.42$\pm$13.89 & 95.91$\pm$0.12 & 42.03$\pm$2.72 & 14.28$\pm$1.04 & 31.16$\pm$3.42 & 93.40$\pm$0.44 & 63.21$\pm$6.84 \\
PN & 33.83$\pm$0.51 & 37.10$\pm$0.71 & 96.96$\pm$0.10 & 26.33$\pm$1.22 & 32.01$\pm$0.61 & 35.37$\pm$0.58 & 95.57$\pm$0.29 & 40.94$\pm$4.49 \\
MDSN & 8.34$\pm$0.46 & 29.04$\pm$6.43 & 97.07$\pm$0.14 & 32.03$\pm$2.29 & 9.44$\pm$0.47 & 38.59$\pm$13.94 & 96.11$\pm$0.12 & 38.91$\pm$3.06 \\
Density Softmax & 8.42$\pm$0.43 & 36.20$\pm$5.78 & 97.03$\pm$0.10 & 32.56$\pm$3.29 & 10.09$\pm$0.40 & 33.59$\pm$4.57 & 95.96$\pm$0.19 & 41.43$\pm$2.25 \\
DAC & 1.42$\pm$0.30 & 18.79$\pm$10.81 & 96.92$\pm$0.10 & 33.89$\pm$2.60 & 2.27$\pm$0.16 & 11.55$\pm$3.48 & 96.08$\pm$0.07 & 40.23$\pm$3.00 \\ \hline
$k$NN-UE (w/o label) & \textbf{1.28$\pm$0.43}& 16.53$\pm$11.45 & 97.09$\pm$0.10 & 30.22$\pm$2.80 & 2.12$\pm$0.36 & 10.00$\pm$6.07 & \textbf{96.12$\pm$0.16} & 37.33$\pm$4.70 \\ 
$k$NN-UE & 1.41$\pm$0.47 & \textbf{10.77$\pm$2.34}\textsuperscript{†} & \textbf{97.18$\pm$0.09}& \textbf{23.83$\pm$1.29}\textsuperscript{†} & \textbf{1.80$\pm$0.37} & \textbf{5.12$\pm$1.47}\textsuperscript{†} & 96.00$\pm$0.22 & \textbf{34.97$\pm$2.48} \\ \hline
\end{tabular}
}
\caption{ECE, MCE, AUROC, and E-AURC results about NLI task on MNLI (In-domain) and SNLI (Out-of-domain) for $\mathrm{DeBERTaV3_{BASE}}$ model.}
\label{tab:nli_debertav3_scores}
\end{table*}

\subsection{Evaluation}
To evaluate the confidence calibration performance, we choose \textit{Expected Calibration Error} (ECE) and \textit{Maximum Calibration Error} (MCE)~\cite{aaai-2015-ece}. For selective prediction, we evaluate \textit{Area Under the Receiver Operator Characteristic curve} (AUROC) and \textit{Excess-Area Under the Risk-Coverage curve} (E-AURC)~\cite{geifman2018biasreduced}. Evaluation metrics computation details are described in Appendix~\ref{sec:details_evaluation_metrics}.
In NER, we performed the evaluation of the UE performance with the flat recombination of the labels and the confidence for all tokens, respectively.

\section{Results}
\label{sec:results}
\subsection{Sentiment Analysis}
\label{sec:results_sa}
In SA, we evaluate the confidence calibration, selective prediction and out-of-distribution detection performance.

\paragraph{Confidence Calibration and Selective Prediction}
First, we present the UE results for sentiment analysis by differentiating the in-domain and out-of-main performance in Table \ref{tab:sentiment_analysis_metrics}. $k$NN-UE consistently outperforms existing methods in terms of ECE, MCE, and E-AURC. In AUROC, LS outperforms in OOD setting, but $k$NN-UE outperforms existing methods in ID setting. Furthermore, the proposed method clearly outperforms DAC that uses neighbor search results for each hidden representation with the additional label term. The lower UE performance than $k$NN-UE in DAC is probably due to the difficulty in optimizing hyperparameters by comprising many layers.

\paragraph{Out-of-Distribution Detection}
Following the previous study~\cite{Tomani-dac-2023}, we carried out the experiments in the out-of-distribution detection task, which determines whether a data instance is in-domain or not. This task is based on the intuition that we want to return predictions with high confidence in ID but with low confidence in predictions in OOD.
We evaluated the out-of-distribution detection performance by using maximum softmax probability as the uncertainty score, and report FPR@95 (the FPR when the TPR is 95\%), AUROC, Area Under the Precision-Recall curve (AUPR)-in and AUPR-out. AUPR-in indicates the AUPR score when ID samples are treated as positive; AUPR-out is vice versa.

Table \ref{tab:sa_ood_evaluation} shows the out-of-distribution detection results when using IMDb/Yelp datasets as ID/OOD, respectively, in $\mathrm{mDeBERTaV3_{BASE}}$ model.
$k$NN-UE consistently shows the out-of-distribution detection performance improvement.
\begin{table}[t]
\scalebox{0.56}{
\begin{tabular}{l|ccccc}
\hline
Methods & FPR@95 (↓) & AUROC (↑) & AUPR-In (↑) & AUPR-Out (↑) \\
\hline
SR & 82.51$\pm$9.49 & 63.18$\pm$5.14 & 69.51$\pm$2.57 & 54.70$\pm$8.48 \\
TS & 83.12$\pm$7.50 & 65.63$\pm$3.64 & 70.99$\pm$2.02 & 56.19$\pm$6.11 \\
LS & 86.88$\pm$4.27 & 62.17$\pm$2.83 & 69.50$\pm$1.51 & 51.38$\pm$3.81 \\
MC Dropout & 87.33$\pm$3.38 & 63.96$\pm$4.09 & 70.13$\pm$2.39 & 53.18$\pm$5.41 \\
SNGP & 81.92$\pm$3.46 & 63.27$\pm$3.07 & 68.83$\pm$2.10 & 55.91$\pm$3.20 \\
PN & 82.84$\pm$5.11 & 67.54$\pm$4.29 & 66.59$\pm$2.45 & 55.32$\pm$5.26 \\
Density Softmax & 87.54$\pm$3.14 & 58.73$\pm$4.33 & 67.34$\pm$2.57 & 49.19$\pm$4.36 \\
DAC & 84.98$\pm$4.19 & 64.65$\pm$6.18 & 70.69$\pm$3.59 & 54.81$\pm$7.29 \\ \hline
$k$NN-UE (w/o label)& 75.87$\pm$2.16 & 70.44$\pm$1.70 & \textbf{74.77$\pm$1.44}\textsuperscript{†} & 63.39$\pm$2.24 \\
$k$NN-UE & \textbf{73.55$\pm$5.01}\textsuperscript{†} & \textbf{71.11$\pm$2.92}\textsuperscript{†} & 73.80$\pm$2.19 & \textbf{65.01$\pm$3.45}\textsuperscript{†} \\
\hline
\end{tabular}
}
\centering
\caption{Out-of-distribution detection results on $\mathrm{mDeBERTaV3_{BASE}}$ model using IMDb/Yelp Polarity as ID/OOD datasets, respectively.}
\label{tab:sa_ood_evaluation}
\end{table}

\subsection{Natural Language Inference}
\label{sec:results_nli}
We show the results of in-domain and out-of-domain UE in NLI task using the DeBERTaV3 model in Table \ref{tab:nli_debertav3_scores}. Similar to Section \ref{sec:results_sa}, $k$NN-UE shows the best UE performance, especially when the label term is included.
\citet{galil2023what} have reported that improving calibration performance does not necessarily lead to the improved selective prediction performance, but our proposed method improves both types of metrics.
On the other hand, the degree of improvement is larger for calibration performance. Specifically, the largest improvement is obtained on SNLI, where $k$NN-UE reduces MCE by more than 31.49 \% compared to SR. Additional experimental results on the Brier score can be found in Appendix \ref{sec:appendix_brier_score}.

\begin{table*}[t!]
\centering
\scalebox{0.64}{
\begin{tabular}{l|ccc|ccc|ccc}
    \hline
    \multicolumn{1}{l|}{Methods} & \multicolumn{3}{c|}{$\mathtt{bn}$ (In-domain)} & \multicolumn{3}{c|}{$\mathtt{nw}$ (Out-of-domain)} & \multicolumn{3}{c}{$\mathtt{tc}$ (Out-of-domain)} \\ \cline{2-10}
      & ECE (↓) & MCE (↓) & E-AURC (↓) & ECE (↓) & MCE (↓) & E-AURC (↓) & ECE (↓) & MCE (↓) & E-AURC (↓) \\ \hline
      SR & 7.79$\pm$0.53 & 50.07$\pm$24.15 & 21.90$\pm$1.31 & 17.05$\pm$0.69 & 37.06$\pm$3.13 & 81.49$\pm$4.17 & 21.20$\pm$2.03 & 42.60$\pm$5.84 & 76.05$\pm$5.72 \\
      TS & 5.34$\pm$0.43 & 75.71$\pm$21.96 & 19.63$\pm$1.22 & 12.76$\pm$0.62 & 26.57$\pm$3.97 & 72.90$\pm$4.72 & 19.69$\pm$0.95 & 47.72$\pm$7.34 & 71.87$\pm$8.83 \\
      LS & 6.46$\pm$0.74 & 50.99$\pm$26.73 & 24.93$\pm$1.19 & 14.78$\pm$0.61 & 30.54$\pm$2.84 & 81.50$\pm$6.98 & 20.99$\pm$2.16 & 65.40$\pm$17.16 & 76.65$\pm$7.33 \\
      MC Dropout & 6.76$\pm$0.64 & 53.13$\pm$26.07 & 19.91$\pm$3.39 & 15.27$\pm$1.01 & 33.60$\pm$4.93 & 77.21$\pm$3.72 & 21.93$\pm$1.63 & 56.56$\pm$12.32 & 75.68$\pm$9.30 \\
      E-NER & 7.98$\pm$0.42 & 61.87$\pm$27.06	& 19.44$\pm$1.81 & 17.42$\pm$0.88 & 40.46$\pm$5.33 & 74.32$\pm$4.47 & 25.42$\pm$2.09 & 59.16$\pm$10.33 & 72.00$\pm$6.57 \\ 
      Density Softmax & 7.32$\pm$0.25 & 59.05$\pm$27.76 & 25.17$\pm$2.63 & 16.10$\pm$0.62 & 44.66$\pm$21.67 & 80.14$\pm$8.50 & 24.40$\pm$1.84 & 62.50$\pm$10.46 & 80.06$\pm$6.27 \\
      DAC & \textbf{1.62$\pm$0.42} & 42.96$\pm$28.25 & 21.47$\pm$2.90 & 7.91$\pm$0.75 & 25.28$\pm$5.15 & 75.24$\pm$2.43 & 14.42$\pm$1.57 & 47.92$\pm$20.98 & 80.72$\pm$8.19 \\ \hline
      $k$NN-UE (w/o label) & 3.37$\pm$0.71 & 33.15$\pm$3.65 & \textbf{17.63$\pm$0.66}\textsuperscript{†} & 8.78$\pm$0.62 & 24.91$\pm$1.81 & \textbf{70.10$\pm$4.03} & 14.61$\pm$0.67 & \textbf{35.26$\pm$7.16}\textsuperscript{†} & \textbf{65.41$\pm$8.11} \\
      $k$NN-UE & 1.78$\pm$0.32 & \textbf{26.02$\pm$13.72} & 20.14$\pm$1.27 & \textbf{7.50$\pm$0.42} & \textbf{16.53$\pm$2.61}\textsuperscript{†} & 74.27$\pm$5.43 & \textbf{14.15$\pm$0.33} & 39.84$\pm$6.02 & 71.81$\pm$9.04 \\ \hline
\end{tabular}
}
\caption{ECE, MCE, and E-AURC results about NER on OntoNotes 5.0 dataset for $\mathrm{mDeBERTaV3_{BASE}}$ model.}
\label{tab:ner_metrics}
\end{table*}

\subsection{Named Entity Recognition}
\label{sec:results_ner}
To evaluate NLP tasks other than simple multi-class classification, we evaluate $k$NN-UE in NER.
Since NER focuses on entities, we use the product of the confidence of the tokens that construct a single entity as the confidence of the entity.

Table \ref{tab:ner_metrics} shows the results of in-domain and out-of-domain UE using the OntoNote 5.0 dataset in $\mathrm{mDeBERTaV3_{BASE}}$.
$k$NN-UE shows the best performance in 4 cases, i.e., ECE or MCE, often resulting in large improvements over SR. On the other hand, E-AURC in NER is consistently better without using the $k$NN-UE label term.\footnote{Label imbalance or large number of class can significantly affect E-AURC on NER when using kNN-UE with the label term. Details are in Appendix~\ref{sec:appendix_behavior_eaurc_ner}.}
E-NER, a recent UE method specifically designed for NER, is close to $k$NN-UE in its entity level selective prediction performance, but the calibration performance is not high.

$k$NN-UE shows good UE performance even when the target domain is relatively far from source domain $\mathtt{bn}$, such as $\mathtt{tc}$.
We have hypothesized that $k$NN-UE might not work if the prediction target is too far from the training data distribution. If the prediction target is too far from the training data, the representation of the prediction from the model will be unreliable when compared to the prediction in the same domain as the training data.
In general, methods based on feature distances assume that they maintain information relevant to the correctness of the prediction~\cite{pmlr-v162-postels22a}.
Our experiments have shown that the problem could be mitigated probably because the domains that the base models do not recognize are limited in the NLP community where there are many strong pre-trained models based on self-supervised learning such as DeBERTaV3.

\begin{figure}[t]
    \includegraphics[width=0.48\textwidth]{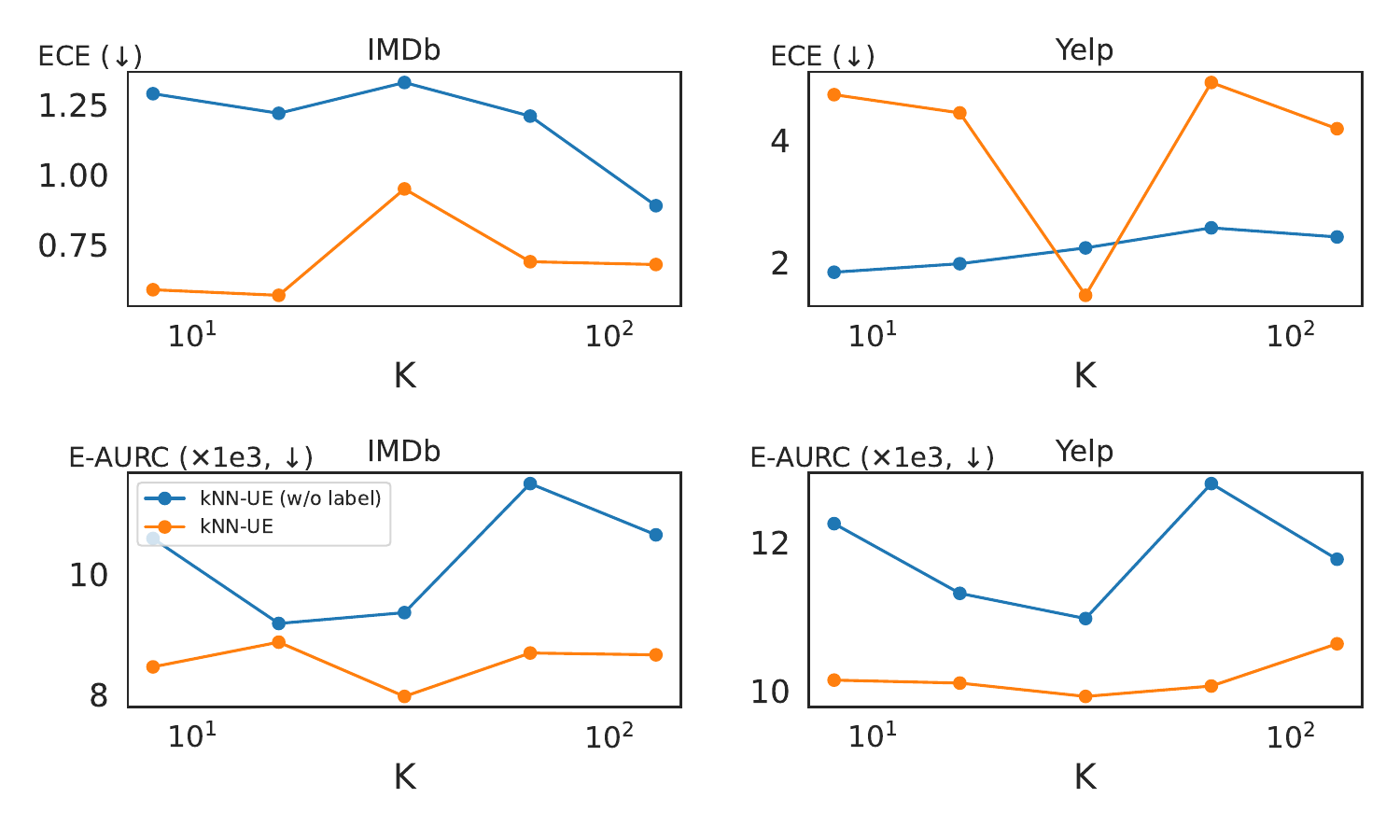}
    \caption{Changes in ECE and E-AURC in SA when changing the number of neighbors of $k$NN-UE. On the x-axis, the parameter $K \in \{8, 16, 32, 64, 128\}$ is represented on a log scale.}
    \label{fig:sentiment_analysis_k_dependency}
\end{figure}

\begin{figure}[t]
    \includegraphics[width=0.48\textwidth]{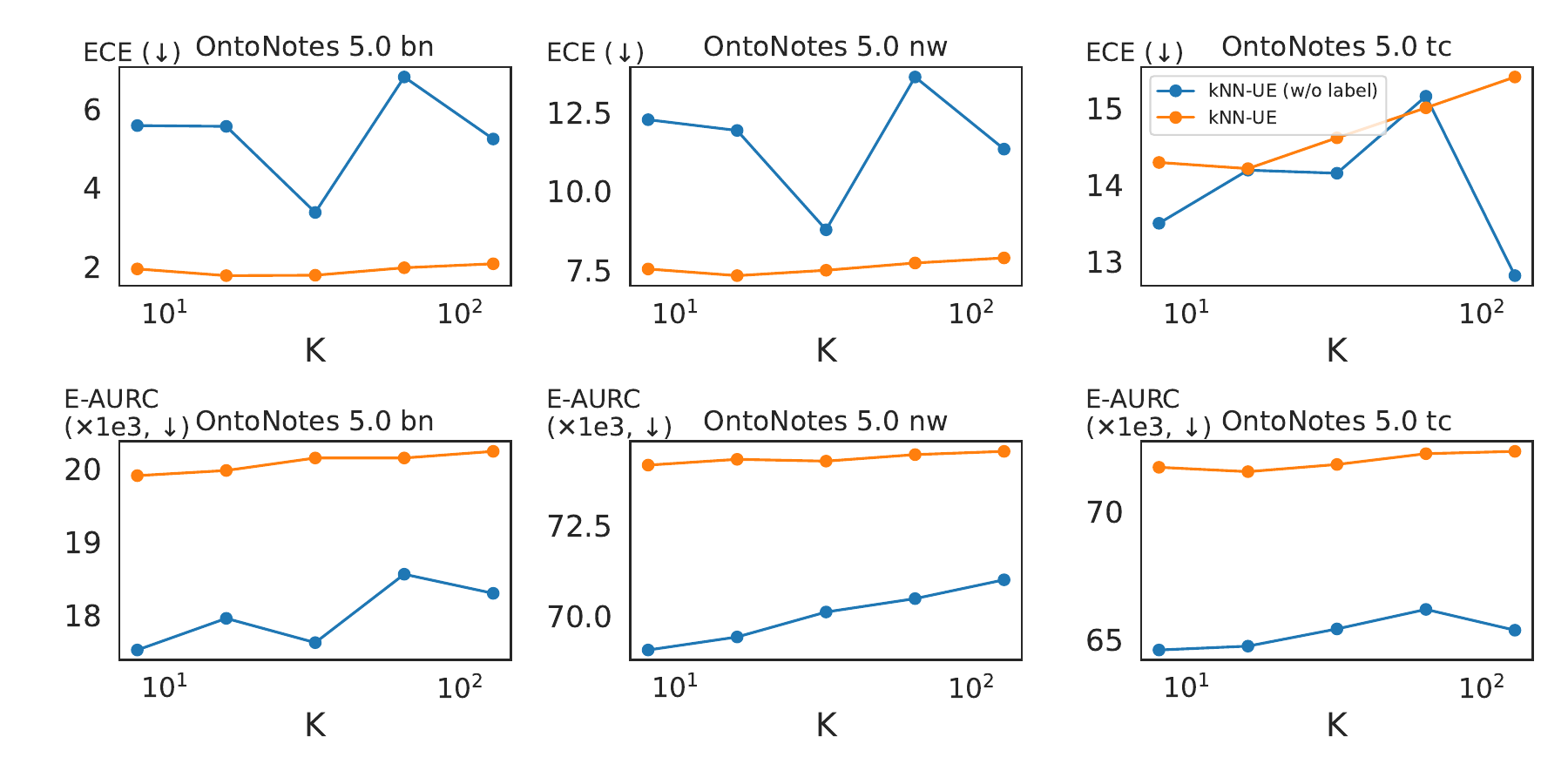}
    \caption{Changes in ECE and E-AURC in NER when changing the number of neighbors of $k$NN-UE. On the x-axis, the parameters $K \in \{8, 16, 32, 64, 128\}$ are represented on a log scale.}
    \label{fig:ner_k_dependency}
\end{figure}

\begin{table}[t]
\scalebox{0.58}{
\begin{tabular}{l|cc|cc}
\hline
 \multicolumn{1}{l|}{Scores} 
 & \multicolumn{2}{c|}{Correct Instances} & \multicolumn{2}{c}{Incorrect Instances} \\ \cline{2-5}
 & $k$NN-UE (w/o label) & $k$NN-UE & $k$NN-UE (w/o label) & $k$NN-UE \\ \hline
$W_{k {\rm{NN}}}$ & 0.49 & 0.50 & 0.41 & 0.27 \\
Confidence & 0.95 & 0.93 & 0.82 & 0.72 \\\hline
\end{tabular}
}
\centering
\caption{Averaged $W_{k {\rm{NN}}}$ and confidence scores with and without label term in $k$NN-UE for correct and incorrect predicted instances when using IMDb as train/validation and Yelp as test, respectively.}
\label{tab:w_knn_scores}
\end{table}

\begin{table}[t]
\scalebox{0.68}{
\begin{tabular}{l|cc}
\hline
Methods & MNLI & OntoNotes 5.0 $\mathtt{bn}$ \\
\hline
SR & 8.41$\pm$0.03 & 2.49$\pm$0.08 \\
TS & 8.42$\pm$0.07 & 2.51$\pm$0.08 \\
LS & 8.44$\pm$0.06 & 2.53$\pm$0.03 \\
MC Dropout & 157.52$\pm$0.51 & 39.81$\pm$0.39 \\
SNGP & 10.58$\pm$2.09 & - \\
PN & 9.11$\pm$0.07 & - \\
MDSN & 9.65$\pm$1.36 & -\\
E-NER & - & 2.51$\pm$0.12 \\
Density Softmax & 8.57$\pm$0.06 & 2.59$\pm$0.05 \\
DAC & 785.15$\pm$6.72 & 183.46$\pm$0.76 \\
$k$NN-UE (w/o label) & 9.05$\pm$0.07 & 4.94$\pm$0.10 \\
$k$NN-UE & 9.08$\pm$0.10 & 4.99$\pm$0.07 \\ \hline
\end{tabular}
}
\centering
\caption{Inference time [s] on MNLI test set and OntoNotes 5.0 $\mathtt{bn}$ test set.}
\label{tab:inference_time_id}
\end{table}

\section{Analysis}
\label{sec:analysis}

\subsection{Impact of Top-\texorpdfstring{$K$}{K}}
\label{sec:impact_of_top_k}
To understand the behavior of $k$NN-UE, we evaluated the performance in UE when changing the number of neighbors $K \in \{8, 16, 32, 64, 128\}$ during $k$NN execution.

Figure \ref{fig:sentiment_analysis_k_dependency} and \ref{fig:ner_k_dependency} show the results for SA and NER, respectively. As is noticeable in NER, the smaller $K$, the better UE tends to be.
These results suggest that our method requires that nearer examples to calibrate confidence, but more distant examples are not important.
When calculating $W_{kNN}(\hat{y})$ in Eq. \ref{eq:knn_ue_weight}, automatically adjusting the importance weights based on the order or distance of the retrieved nearest neighbors could further improve UE performance.
Similar experimental and theoretical analysis of out-of-distribution detection using only $k$NN distance also suggests that using $k$-th example is preferable~\cite{pmlr-v162-sun22d}.
Providing a similar theoretical justification for our $k$NN-UE is an interesting future direction.

\begin{table*}[t!]
\centering
\scalebox{0.68}{
\begin{tabular}{l|cccc|cccc}
\hline
 & \multicolumn{4}{c|}{OntoNotes 5.0 $\mathtt{bn}$ (In-domain)} & \multicolumn{4}{c}{OntoNotes 5.0 $\mathtt{nw}$ (Out-of-domain)} \\ \cline{2-9}
\multicolumn{1}{l|}{Methods} & ECE (↓) & MCE (↓) & E-AURC (↓) & time [s] & ECE (↓) & MCE (↓) & E-AURC (↓) & time [s] \\ \hline
SR                  & 7.79$\pm$0.53     & 50.07$\pm$24.15   & 21.90$\pm$1.31     & 2.49$\pm$0.08 & 17.05$\pm$0.69    & 37.06$\pm$3.13    & 81.49$\pm$4.17     & 5.75$\pm$0.27 \\
$k$NN-UE (w/o label)  & 3.37$\pm$0.71     & 33.15$\pm$3.65    & 17.63$\pm$0.66     & 4.94$\pm$0.10 & 8.78$\pm$0.62     & 24.91$\pm$1.81    & 70.10$\pm$4.03     & 10.36$\pm$0.21 \\
$k$NN-UE & 1.78$\pm$0.32     & 26.02$\pm$13.72   & 20.14$\pm$1.27     & 4.99$\pm$0.07 & 7.50$\pm$0.42     & 16.53$\pm$2.61    & 74.27$\pm$5.43     & 10.48$\pm$0.12 \\ \hline
+ PQ & 1.96$\pm$0.31     & 31.33$\pm$18.74   & 20.23$\pm$1.27     & 3.32$\pm$0.05  & 7.57$\pm$0.45     & 16.43$\pm$2.73    & 74.38$\pm$5.36     & 7.23$\pm$0.16\\
+ IVF & 1.92$\pm$0.31     & 28.55$\pm$11.24     & 20.13$\pm$1.22 & 3.31$\pm$0.06  & 7.60$\pm$0.41 & 17.12$\pm$2.35  & 74.34$\pm$5.35  & 7.33$\pm$0.21 \\ 
+ DR  & 2.14$\pm$0.37  & 33.52$\pm$10.84  & 20.12$\pm$1.26	  & 2.87$\pm$0.04 & 8.08$\pm$0.53  & 24.03$\pm$5.46  & 74.50$\pm$5.42  & 6.20$\pm$0.20 \\ \hline
\end{tabular}
}
\caption{ECE, MCE, E-AURC and inference time results about NER on OntoNotes 5.0 $\mathtt{bn}$ (In-domain) and OntoNotes 5.0 $\mathtt{nw}$ (Out-of-domain) for $\mathrm{mDeBERTaV3_{BASE}}$ model when applied PQ, IVF, and dimension reduction sequentially. DR indicates dimension reduction. For comparison, we also present the results when dimension reduction is only applied to $k$NN-UE.}
\label{tab:combination_scores}
\end{table*}

\begin{table}[t]
\centering
\scalebox{0.70}{
\begin{tabular}{l|cc}
\hline
\multicolumn{1}{l|}{Methods} & OntoNotes 5.0 $\mathtt{bn}$ & OntoNotes 5.0 $\mathtt{nw}$ \\ \hline
$k$NN-UE & 100.0 & 100.0 \\ \hline
+ PQ & 21.30 & 51.68  \\
+ IVF  & 18.60 & 11.04 \\
+ DR & 0.02 & 0.04  \\ \hline
Only DR & 43.98 & 20.35 \\ \hline
\end{tabular}
}
\caption{Coverages when PQ, clustering, and PCA are applied sequentially to the example indices obtained by default $k$NN-UE. Results when applying dimension reduction by PCA individually are also presented for reference.}
\label{tab:combination_coverages}
\end{table}

\subsection{Importance of Label Term in $W_{k {\rm{NN}}}$}
\label{sec:importance_label_term_wknn}
We analyze the impact of the label term Eq. \ref{eq:knn_ue_weight} on the $k$NN-UE confidence computation. We have shown that the UE performance is improved in several experiments. However, it is not obvious whether the improvement in UE performance is due to the reduction in $W_{k {\rm{NN}}}$ primarily caused by adding the label term. Therefore, we examined the $W_{k {\rm{NN}}}$ values for correctly and incorrectly predicted instances in both the absence and presence of the label term in $k$NN-UE. Table~\ref{tab:w_knn_scores} shows the distance terms, label terms and $W_{k {\rm{NN}}}$ results for each case. If the predictions are correct, the growth of $W_{k {\rm{NN}}}$ due to the label term is limited. On the other hand, $k$NN-UE with label term remarkably reduce $W_{k {\rm{NN}}}$ leading to the reduced confidence when the predictions are incorrect. This result suggests that the improvement of the evaluation metrics in $k$NN-UE with label term is not achieved by increasing the confidence when the prediction is correct, but by appropriately reducing the confidence when the prediction is incorrect.

\subsection{Impact of Efficient Nearest Neighbor Search Techniques}
\label{sec:impact_efficient_ann}
We investigate the inference time and UE performance when applying approximate nearest neighbor search techniques and dimension reduction when executing $k$NN search in $k$NN-UE as a real world application.
As shown in Table~\ref{tab:inference_time_id},\footnote{Other results can be found in Appendix~\ref{sec:appendix_inference_time_results}.} in the \textit{sequence labeling} based NER, which requires executing $k$NN searches per token, it takes twice as much inference time as SR.\footnote{Inference times do not increase as dramatically as $k$-Nearest Neighbor Language Model~\cite{Khandelwal2020Generalization} because $k$NN can be executed in parallel for both classification and NER.}
On the other hand, in $k$-Nearest Neighbor Language Model ($k$NN-LM)~\cite{Khandelwal2020Generalization}, dimension reduction and approximate $k$NN search techniques are effective to improve inference speed while maintaining perplexity in text generation~\cite {he-etal-2021-efficient, pmlr-v202-xu23a}.
Therefore, inspired by these works for faster $k$NN-LM, we investigate how the approximate nearest neighbor search techniques, such as Product Quantization~\cite{product_quantization_2011} (PQ), Inverted File (IVF) clustering and dimension reduction affect the UE and inference speed of our $k$NN-UE. Description of approximate nearest neighbor search techniques and detailed discussion when each method is individually applied to $k$NN-UE are in Appendix~\ref{sec:appenedix_each_efficient_technique}.

\paragraph{Results of Combination of PQ, IVF and Dimension Reduction}
We evaluate the UE performance and inference speed when applying PQ, IVF and dimension reduction are applied. Table \ref{tab:combination_scores} shows the results on OntoNotes 5.0 $\mathtt{bn}$ and $\mathtt{nw}$ test sets as ID/OOD, respectively. The detailed discussion when changing the parameters of PQ, clustering and dimension reduction are shown in Appendix \ref{sec:appenedix_each_efficient_technique}. 
We can see that ECE and MCE are degraded when PQ, IVF and dimension reduction by PCA are applied simultaneously to $k$NN-UE.\footnote{Distance recomputation does not mitigate this behavior, see Appendix~\ref{sec:appendix_distance_recomputaton}.}
On the other hand, our results show that applying them appropriately such as combining PQ with IVF improve inference time with mitigating the degradation in UE performance (The results with the parameters for PQ or IVF can be found in Appendix~\ref{sec:appendix_analysis_product_quantization} or~\ref{sec:appendix_analysis_clustering}).
To deepen our understanding of the above changes in the behavior of the uncertainty performance due to appling of approximate $k$NN search techniques or dimension reduction in $k$NN-UE, we calculated the coverage that how much the indices obtained when using the default exhaustive search are covered when applying PQ, clustering, and dimension reduction sequentially.

Table~\ref{tab:combination_coverages} shows the coverages on OntoNotes 5.0 $\mathtt{bn}$ and $\mathtt{nw}$ as ID/OOD settings, respectively.
We can see that applying PQ, clustering, and PCA simultaneously hardly covers any of the indices from the default $k$NN-UE.
It is assumed that applying PQ and PCA in the same time leads to coarse distance computation in a single subvector, which would correspondingly degrade the UE performance in $k$NN-UE.
Actually, the experimental results in Table \ref{tab:dimension_reduction_scores} in Appendix \ref{sec:appendix_analysis_dimension_reduction} suggest that excessive dimension reduction in distance computation could have a negative impact on the UE performance.
On the other hand, if combined with PQ and IVF, or applied PCA individually, some of the ground-truth nearest neighbor examples still exist.






\section{Conclusion}
In this paper, we proposed $k$NN-UE, which estimates uncertainty by using the distance to neighbors and labels of neighbors. The experimental results showed that our method showed higher UE performance than existing UE methods in SA, NLI and NER.
Furthermore, our analysis of correctly and incorrectly predicted instances suggests that the improvement in $k$NN-UE is largely due to the reduction in confidence on incorrect instances.
In addition, we investigated the effects of efficient neighbor search techniques in $k$NN-UE to address the degradation of the inference speed in token-level tasks such as NER.
As a result, we found that product quantization, clustering, or dimension reduction improves inference speed without degrading the UE much more, unless combining all of them simultaneously.

\section{Limitations}
In this study, we focused only on the classification-based tasks.
On the other hand, taking advantage of the recent growth of Large Language Models, UE in text generation is also attracting attention~\cite{yoshikawa-okazaki-2023-selective, fadeeva-etal-2023-lm, lin2024generating}.
Therefore, to investigate the effectiveness of $k$NN-UE in text generation tasks is an interesting direction for future research. Not only that, our proposed method is applicable to more tasks such as image classification.

Furthermore, although $k$NN-UE only used the representation of the last layer of the base model, exploring for an appropriate representation for UE is a future challenge.
Also, to investigate the relationship between the representation quality and in- and out-of-domain UE performance when using smaller pretrained encoders than DeBERTa, such as BERT~\cite{devlin-etal-2019-bert} and RoBERTa~\cite{roberta-2019} is an interesting direction.

Finally, we used ECE and MCE to measure calibration performance. On the other hand, it may be more appropriate to use other metrics to measure calibration performance when the dataset with multiple annotations including human disagreement is available~\cite{baan-etal-2022-stop}, where it may be similar to the label disagreement in similar output representations.
In Section ~\ref{sec:importance_label_term_wknn}, we showed that our $k$NN-UE with the label term improves in the direction we expected: it reduces confidence much more when the predictions are inaccurate.
However, measuring calibration performance on a variety of data with multiple annotations may provide a more interesting insight into the behavior of our proposed method.

\section*{Ethical Considerations}
In this study, we used existing datasets that have cleared ethical issues following policies of published conferences. Therefore, they do not introduce any ethical problems. On the other hand, we have an ethical consideration about UE. Specifically, decision support systems with machine learning algorithms do not necessarily have a positive effect on performance. \citet{2021-translational-psychiatry} showed that collaboration with machine learning models does not significantly improve clinician's treatment selection performance, and that performance is significantly degraded due to the presentation of incorrect recommendations. This problem is expected to remain even if UE methods are applied to machine learning models. In addition, introducing UE methods could conversely lead humans to give overconfidence in machine learning models, resulting in performance degradation.

\section*{Acknowledgements}
The authors also acknowledge the Nara Institute of Science and Technology's HPC resources made available for conducting the research reported in this paper.

\bibliography{custom}

\appendix

\section{Dataset Statistics}
\label{sec:appendix_dataset_statistics}
The dataset statistics in our study is shown in Table~\ref{tab:dataset_statistics}.

\begin{table}[h]
\centering
\scalebox{0.68}{
\begin{tabular}{l|l|l|rrr}
\hline
Tasks & Datasets & $N_{class}$ & Train & Val & Test \\ \hline
SA & \textbf{IMDb} & 2 & 25,000 & 12,500 & 12,500 \\ 
 & Yelp & 2 & - & - & 19,000 \\ \hline
NLI & \textbf{MNLI} & 3 & 392,702 & 4,907 & 4,908 \\ 
 & SNLI & 3 & - & - & 9,824 \\ \hline
NER & \textbf{OntoNotes 5.0 ($\mathtt{bn}$)} & 37 & 10,683 & 1,295 & 1,357 \\
 & OntoNotes 5.0 ($\mathtt{nw}$) & 37 & - & - & 2,327 \\ 
 & OntoNotes 5.0 ($\mathtt{tc}$) & 37 & - & - & 1,366 \\ \hline
\end{tabular}
}
\caption{Dataset Statistics. Bolds indicate In-domain.}
\label{tab:dataset_statistics}
\end{table}

\section{Training Settings for Density Estimator in Density Softmax}
\label{sec:appendix_training_density_estimator}
In Density Softmax~\cite{pmlr-v235-bui24a}, we use RealNVP~\cite{dinh2017density} as the density estimator, which has two coupling structures. Table \ref{tab:hyperparameters_density_softmax} shows the hyperparameters for training RealNVP as the density estimator in Density Softmax.

\begin{table}[ht]
\centering
\scalebox{0.68}{
\begin{tabular}{l|c}
\hline
Hyperparameters & Values \\ \hline 
learning rate & 1e-4  \\ 
optimizer & AdamW~\cite{adamw-2019} \\ 
early stopping patient & 5 \\
\makecell[l]{number of coupling layers} & 4 \\
hidden units & 16 \\ \hline
\end{tabular}
}
\caption{Hyperparameters for RealNVP in Density Softmax.}
\label{tab:hyperparameters_density_softmax}
\end{table}

\section{Details of Baselines}
\label{sec:appendix_details_baselines}
\paragraph{Softmax Response (SR)} is a trivial baseline, which treats the maximum score from output of the base model's softmax layer as the confidence~\cite{1995SR}.

\paragraph{Temperature Scaling (TS)} is a calibration technique by which the logits are divided by a temperature parameter $T$ before applying the softmax function~\cite{pmlr-v70-guo17a}. We optimized $T$ by L-BFGS on validation set loss.

\paragraph{Label Smoothing (LS)} is the calibration and generalization technique by introducing a small degree of uncertainty $\epsilon$ in the target labels during training~\cite{Miller1996AGO, pereyra2017regularizing}. 
In LS, we optimized $\epsilon \in \{0.01, 0.05, 0.1, 0.2, 0.3\}$ by using validation set accuracy when SA and NLI, and validation set $F_{1}$ when NER.

\paragraph{MC Dropout} is an UE technique by $M$ times stochastic inferences with activating dropout~\cite{pmlr-v48-gal16}. In our experiments, we set $M = 20$ for all evaluations, and the dropout rate is 0.1.

\paragraph{Spectral-Normalized Gaussian Process (SNGP)} uses spectral normalization of the weights for distance-preserving representation and Gaussian Processes in the output layer for estimating uncertainty~\cite{NEURIPS2020_SNGP}.

\paragraph{Posterior Networks (PN)} is one of the methods in the Evidential Deep Learning (EDL) framework~\cite{NEURIPS2018_edl} that assumes a probability distribution for class probabilities~\cite{posterior_network_NEURIPS2020}, which uses normalizing flow~\cite{pmlr-v37-rezende15} to estimate the density of each class in the latent space.

\paragraph{Mahalanobis Distance with Spectral-Normalized Network (MDSN)} is a Mahalanobis distance based UE method that benefits from by spectral normalization of the weights~\cite{vazhentsev-etal-2022-uncertainty}, similar to SNGP.

\paragraph{E-NER} applies EDL framework for NER by introducing uncertainty-guided loss terms~\cite{zhang-etal-2023-ener}.



\section{Details of Evaluation Metrics}
\label{sec:details_evaluation_metrics}
\paragraph{Expected Calibration Error (ECE)}
ECE \cite{aaai-2015-ece} quantifies the difference between the accuracy and confidence of a model. Formally, ECE is expressed as:
\begin{equation}
    \text{ECE} = \sum_{b=1}^B \frac{|\mathcal{D}_b|}{n} \left \lvert \text{acc}( \mathcal{D}_b) - \text{conf}( \mathcal{D}_b) \right \rvert 
\end{equation}
where $B$ is the number of confidence interval bins, $\mathcal{D}_b$ denotes the set of examples with predicted confidence scores in the $b$-th bin, $n$ is the total number of examples, $\text{acc}(\mathcal{D}_b)$ is the accuracy of the model on the examples in $\mathcal{D}_b$, and $\text{conf}(\mathcal{D}_b)$ is the average confidence of the model on the examples in $\mathcal{D}_b$. In this study, we use $B=10$.

\paragraph{Maximum Calibration Error (MCE)}
MCE, as detailed by~\citet{aaai-2015-ece} measures the maximum difference between the model's accuracy and the confidence across variousb confidence levels. MCE is defined as:
\begin{equation}
    \text{MCE} = \max_{b=1}^B \left\lvert \text{acc}(\mathcal{D}_b) - \text{conf}(\mathcal{D}_b) \right\rvert,
\end{equation}
A lower MCE means that there is a small risk that the confidence of the model's prediction will deviate greatly from the actual correct answer. In this study, we use $B=10$, same as ECE.

\paragraph{Area Under the Risk-Coverage curve (AURC)}
The AURC is the area of the risk-coverage curve when the confidence levels of the forecasts corresponding to the $N$ data points are sorted in descending order. The larger the area, the lower the error rate corresponding to a higher confidence level, which means that the output confidence level is more appropriate. Formally, AURC is defined as:
\begin{equation}
    \text{AURC} = \sum_{n=1}^N \frac{\sum_{j=1}^{n} g(x_{j})}{i \times N}
\end{equation}
where $g(x)$ returns 1 if the prediction is wrong and 0 otherwise. 

\paragraph{Excess-Area Under the Risk-Coverage curve (E-AURC)}
E-AURC \cite{geifman2018biasreduced} is a measure of the AURC score normalized by the smallest risk-coverage curve area $\text{AURC}^\star \approx \hat{r} + (1 - \hat{r}) \mathrm{ln} (1 - \hat{r})$, where $\hat{r}$ is the error rate of the model. The reason for normalizing the AURC is that the AURC depends on the predictive performance of the model and allows for performance comparisons of confidence across different models and training methods. E-AURC is defined as:
\begin{align}
    \text{E-AURC} &= \text{AURC} - \text{AURC}^\star
\end{align}
E-AURC scores are reported with multiplying by 1,000 due to visibility.

\section{Additional Results on the Brier score}
\label{sec:appendix_brier_score}
The Brier score is a widely used metric in UE community for evaluating the probabilistic predictions. The metric measures the mean squared difference between the predicted probability assigned to the predicted label and the actual outcome. This evaluation serves as a holistic assessment of model performance, reflecting both fit and calibration, in the following formula: 

\begin{align}
    \text{Brier score} &= \frac{1}{N} \sum_{n=1}^{N} (p_{n} - o_{n}),
\end{align}

where $p_{n}$ is the predicted probability assigned to the prediction, and $o_{n}$ is the actual outcome.
Table \ref{tab:brier_scores} shows the results on the Brier score. These results indicate $k$NN-UE improves calibration performance more prominently than other methods while maintaining prediction performance.

\begin{table}[t]
\scalebox{0.56}{
\begin{tabular}{l|ccccc}
\hline
Methods & \multicolumn{2}{c}{SA} & \multicolumn{2}{c}{NLI}  \\
 & IMDb & Yelp & MNLI & SNLI \\
\hline
SR & 5.00$\pm$0.27 & 5.83$\pm$0.98 & 9.50$\pm$0.40 & 11.02$\pm$0.41 \\
TS & 5.09$\pm$0.42 & 6.67$\pm$1.36 & 8.31$\pm$0.25 & 9.60$\pm$0.21 \\
LS & 4.64$\pm$0.23 & 5.16$\pm$0.92 & 8.73$\pm$0.23 & 10.18$\pm$0.17 \\
MC Dropout & 4.88$\pm$0.21 & 5.45$\pm$0.55 & 9.33$\pm$0.36 & 11.00$\pm$0.28 \\
SNGP & 4.78$\pm$0.15 & 5.99$\pm$0.39 & 12.25$\pm$5.38 & 13.45$\pm$4.57 \\
PN & 10.31$\pm$0.28 & 11.16$\pm$0.22 & 20.76$\pm$0.32 & 21.11$\pm$0.42 \\
Density Softmax & 4.82$\pm$0.18 & 6.05$\pm$0.38 & 9.60$\pm$0.34 & 11.28$\pm$0.41 \\
DAC & 4.44$\pm$0.33 & 5.44$\pm$0.71 & 8.21$\pm$0.25 & 9.55$\pm$0.35 \\ \hline
kNN-UE (w/o label) & 4.37$\pm$0.16 & 5.10$\pm$0.12 & 8.15$\pm$0.15 & 9.52$\pm$0.32 \\
kNN-UE & \textbf{4.21$\pm$0.14} & \textbf{5.02$\pm$0.42} & \textbf{8.07$\pm$0.18} & \textbf{9.44$\pm$0.28} \\ \hline
\end{tabular}
}
\centering
\caption{Brier score results using IMDb/Yelp and MNLI/SNLI as ID/OOD datasets, respectively.}
\label{tab:brier_scores}
\end{table}

\section{The impact of \texorpdfstring{$k$}{k}NN-UE with label term in NER on E-AURC}
\label{sec:appendix_behavior_eaurc_ner}
NER tasks are often in label imbalanced settings, where the "O" label is typically much more than other entity-related labels.
Additionally, in the OntoNotes 5.0 dataset, the number of labels is 37, as shown in Table~\ref{tab:dataset_statistics}, which is significantly higher than in SA and NLI tasks.
As a result, compared to SA and NLI, neighbor labels will contain much more different labels from the predicted label.
The presence of many other labels in the neighbors that are different from the predicted label can lead to excessively low confidence in $k$NN-UE using the label term, even though the prediction is correct because $S(\hat{y})$ in the label term becomes lower in NER.
The impact of that bias for calibration errors, such as ECE and MCE, will be limited.
However, low confidence in accurate prediction reduces the coverage much in E-AURC, leading to a degradation in E-AURC.

For example, assume that in a test data set of 6 examples for 3 classes classification, the predictions for the first 3 examples are incorrect and the latter 3 examples are correct.
In case A, we assume that prediction confidences are [[0.25, 0.25, 0.5], [0.25, 0.25, 0.5], [0.25, 0.25, 0.5], \textbf{[0.25, 0.25, 0.5]}, [0.02, 0.02, 0.96], [0.01, 0.01, 0.98]].
In case B, we assume that prediction confidences are [[0.25, 0.25, 0.5], [0.25, 0.25, 0.5], [0.25, 0.25, 0.5], \textbf{[0.275, 0.275, 0.45]}, [0.02, 0.02, 0.96], [0.01, 0.01, 0.98]].
In these settings, the ECE and E-AURC for each case are shown in Table~\ref{tab:behavior_ner_ece_eaurc_toy}.
These scores indicate that E-AURC is strongly penalized when the confidence in a correct prediction is lower than the confidence in an incorrect prediction.

\begin{table}[t]
\scalebox{0.85}{
\begin{tabular}{l|cc}
\hline
Case & ECE (↓) & E-AURC (↓) \\
\hline
A & 17.67 & 18.80 \\
B & 16.83 & 121.57 \\ \hline
\end{tabular}
}
\centering
\caption{ECE and E-AURC in two toy cases of Appendix~\ref{sec:appendix_behavior_eaurc_ner}.}
\label{tab:behavior_ner_ece_eaurc_toy}
\end{table}

\section{Inference Time Full Results}
\label{sec:appendix_inference_time_results}
We show the inference time full results on out-of-domain test sets in Table~\ref{tab:inference_time_ood}.

\begin{table}[t]
\scalebox{0.68}{
\begin{tabular}{l|cc}
\hline
Methods & SNLI & OntoNotes 5.0 $\mathtt{nw}$ \\
\hline
SR & 21.59$\pm$0.76 & 5.75$\pm$0.27 \\
TS & 21.64$\pm$0.07 & 5.79$\pm$0.17 \\
LS & 21.70$\pm$0.07 & 5.80$\pm$0.19 \\
MC Dropout & 396.86$\pm$1.10 & 101.98$\pm$0.83 \\
SNGP & 24.59$\pm$0.08 & - \\
PN & 23.26$\pm$0.05 & - \\
MDSN & 23.39$\pm$0.85 & -\\
E-NER & - & 5.78$\pm$0.61 \\
Density Softmax & 22.02$\pm$0.05 & 6.02$\pm$0.07 \\
DAC & 2346.62$\pm$36.06 & 326.00$\pm$1.41 \\ \hline
$k$NN-UE (w/o label) & 23.02$\pm$0.04 & 10.36$\pm$0.21 \\
$k$NN-UE & 23.07$\pm$0.05 & 10.48$\pm$0.12 \\ \hline
\end{tabular}
}
\centering
\caption{Inference time [s] on SNLI test set and OntoNotes 5.0 $\mathtt{nw}$ test set.}
\label{tab:inference_time_ood}
\end{table}

\section{Inference Time Results When Changing -\texorpdfstring{$K$}{K}}
\label{sec:appendix_inference_time_results_changing_k}
To estimate whether the inference time changes significantly when changing Top-$K$ in kNN search, we investigated the inference time when changing $K$ on the IMDb test set. Table~\ref{tab:inference_time_changing_k} shows that the inference time remains almost the same when changing $K$ in the range of 8 to 128.

\begin{table}[t!]
\scalebox{0.68}{
\begin{tabular}{l|c}
\hline
Methods & Inference time [s] \\
\hline
SR & 121.56$\pm$0.12 \\ \hline
$k$NN-UE ($K$=8) & 128.98$\pm$0.11 \\
$k$NN-UE ($K$=16) & 128.97$\pm$0.13 \\
$k$NN-UE ($K$=32) & 128.54$\pm$0.16 \\
$k$NN-UE ($K$=64) & 129.16$\pm$0.16 \\
$k$NN-UE ($K$=128) & 128.39$\pm$0.20 \\ \hline
\end{tabular}
}
\centering
\caption{Inference time [s] on IMDb test set when changing $K$ in $k$NN-UE.}
\label{tab:inference_time_changing_k}
\end{table}

\section{Each Result of Product Quantization, Clustering, and Dimension Reduction}
\label{sec:appenedix_each_efficient_technique}



\subsection{Product Quantization}
\label{sec:appendix_analysis_product_quantization}
(PQ)~\cite{product_quantization_2011} is a data compression technique based on vector quantization. In PQ, a $D$-dimensional representation is divided into $N\textsubscript{sub}$ subvectors and quantized by performing $k$-means clustering on the vectors in each subspace. Vector quantization can significantly reduce the amount of memory occupied by vectors.\footnote{For example, raw datastore in kNN-UE is 636MB on OntoNotes 5.0 $\mathtt{bn}$, but PQ reduces it to 10MB.} In addition, by calculating the distance between compressed PQ codes, we can efficiently calculate the estimated value of the original Euclidean distance.

We evaluated UE performance and inference time when the number of clusters in the codebook was fixed at 32, and the number of subvectors was changed to $N\textsubscript{sub} \in \{16, 32, 64\}$ (In Table~\ref{tab:combination_scores} and ~\ref{tab:combination_coverages}, PQ was performed with $N\textsubscript{sub} = 32$).\par
Table \ref{tab:product_quantization_scores} shows the UE performance and inference time results in different $N\textsubscript{sub}$.
In ECE and E-AURC, there are almost no degradation in UE performance due to PQ. On the other hand, in MCE in ID setting, the UE performance consistently degrades.
Furthermore, compared to $k$NN-UE among different $N\textsubscript{sub}$, the larger $N\textsubscript{sub}$, the better the UE performance tends to improve, but the inference time increases.

The larger $N\textsubscript{sub}$ is, the more time is required for inference but the UE performance improves.
We assumed that these results are derived from the decrease in quantization error over the vector with PQ with larger $N\textsubscript{sub}$ because each subvector is divided into smaller subspaces and the quantization is performed for each subspace.
On the other hand, an increase in $N\textsubscript{sub}$ requires additional distance computations etc., then more inference time.

\begin{table}[t]
\centering
\scalebox{0.56}{
\begin{tabular}{l|cccc}
\hline
\multicolumn{1}{l|}{Methods} & ECE (↓) & MCE (↓) & E-AURC (↓) & time [s] \\ \hline
& \multicolumn{4}{c}{OntoNotes 5.0 $\mathtt{bn}$ (In-domain)} \\ 
SR                          & 7.79$\pm$0.53     & 50.07$\pm$24.15   & 21.90$\pm$1.31     & 2.49$\pm$0.08      \\
$k$NN-UE (w/o label)          & 3.37$\pm$0.71     & 33.15$\pm$3.65    & 17.63$\pm$0.66     & 4.94$\pm$0.10      \\
$k$NN-UE                      & 1.78$\pm$0.32     & 26.02$\pm$13.72   & 20.14$\pm$1.27     & 4.99$\pm$0.07      \\ \hline
$k$NN-UE ($N\textsubscript{sub} = 16$) & 1.90$\pm$0.27     & 31.18$\pm$11.17   & 20.16$\pm$1.12     & 3.27$\pm$0.06      \\
$k$NN-UE ($N\textsubscript{sub} = 32$) & 1.96$\pm$0.31     & 31.33$\pm$18.74   & 20.23$\pm$1.27     & 3.32$\pm$0.05      \\
$k$NN-UE ($N\textsubscript{sub} = 64$) & 1.88$\pm$0.34     & 31.06$\pm$16.36   & 20.16$\pm$1.23     & 4.11$\pm$0.11      \\ \hline
& \multicolumn{4}{c}{OntoNotes 5.0 $\mathtt{nw}$ (Out-of-domain)} \\ 
SR                          & 17.05$\pm$0.69    & 37.06$\pm$3.13    & 81.49$\pm$4.17     & 5.75$\pm$0.27      \\
$k$NN-UE (w/o label)          & 8.78$\pm$0.62     & 24.91$\pm$1.81    & 70.10$\pm$4.03     & 10.36$\pm$0.21     \\
$k$NN-UE                      & 7.50$\pm$0.42     & 16.53$\pm$2.61    & 74.27$\pm$5.43     & 10.48$\pm$0.12     \\ \hline
$k$NN-UE ($N\textsubscript{sub} = 16$) & 7.66$\pm$0.48     & 17.07$\pm$3.81    & 74.47$\pm$5.53     & 7.22$\pm$0.19      \\
$k$NN-UE ($N\textsubscript{sub} = 32$) & 7.57$\pm$0.45     & 16.43$\pm$2.73    & 74.38$\pm$5.36     & 7.23$\pm$0.16      \\
$k$NN-UE ($N\textsubscript{sub} = 64$) & 7.57$\pm$0.44     & 16.38$\pm$2.66    & 74.35$\pm$5.49     & 8.90$\pm$0.18      \\ \hline
\end{tabular}
}
\caption{ECE, MCE, E-AURC and inference time results about NER  on OntoNotes 5.0 $\mathtt{bn}$ (In-domain) and OntoNotes 5.0 $\mathtt{nw}$ (Out-of-domain) for $\mathrm{mDeBERTaV3_{BASE}}$ model when applied PQ in different $N\textsubscript{sub}$.}
\label{tab:product_quantization_scores}
\end{table}

\subsection{Clustering}
\label{sec:appendix_analysis_clustering}
The original $k$NN-LM uses an inverted file index (IVF) technique that speeds up the search by dividing the representation into $N\textsubscript{list} $ clusters by $k$-means and searching for neighbors based on $N\textsubscript{probe}$ centroids. In this study, we evaluate the UE performance and inference speed when the number of clusters $N\textsubscript{list} = 100$.
In this study, we evaluate the UE performance and inference speed when the number of clusters $N\textsubscript{list} = 100$ and applying PQ with $N\textsubscript{sub} = 32$ are fixed and the number of cluster centroids to search changes$N\textsubscript{probe} \in \{8, 16, 32, 64\}$ (In Table~\ref{tab:combination_scores} and ~\ref{tab:combination_coverages}, IVF was performed with $N\textsubscript{probe} = 32$).\par

Table \ref{tab:clustering_scores} shows the performance of UE when changing $N\textsubscript{probe}$ in ID and OOD settings using OntoNotes 5.0. In ECE, scores are slightly reduced for ID, but only slightly worse for OOD; MCE also shows degradation for ID but little for OOD, and even improves when $N\textsubscript{probe} = 8$; E-AURC shows almost no change in scores when $N\textsubscript{probe}$ is changed for both ID and OOD. In terms of inference time, the larger $N\textsubscript{probe}$, the longer it takes.
We derive the improvement in MCE when increasing $N\textsubscript{probe}$ in ID setting from the fact that more clusters are targeted, making it possible to cover ground-truth nearest neighbor examples.
On the other hand, the tendency of slight decrease when increasing $N\textsubscript{probe}$ in OOD setting may comes from the reliability of the vector, similar to the discussion in Section~\ref{sec:results_ner}.

In addition, Taken together with the results in Table~\ref{tab:combination_scores} in Section ~\ref{sec:impact_efficient_ann}, we can see that the degradation of the UE performance can be mitigated with improvement latency when applying PQ and IVF with lower $N\textsubscript{probe}$, compared to applying PQ, IVF and PCA simultaneously.

\begin{table}[t]
\centering
\scalebox{0.56}{
\begin{tabular}{l|cccc}
\hline
\multicolumn{1}{l|}{Methods} & ECE (↓) & MCE (↓) & E-AURC (↓) & time [s] \\ \hline
& \multicolumn{4}{c}{OntoNotes 5.0 $\mathtt{bn}$ (In-domain)} \\ 
SR                  & 7.79$\pm$0.53     & 50.07$\pm$24.15     & 21.90$\pm$1.31     & 2.49$\pm$0.08      \\
$k$NN-UE (w/o label)  & 3.37$\pm$0.71     & 33.15$\pm$3.65      & 17.63$\pm$0.66     & 4.94$\pm$0.10      \\
$k$NN-UE              & 1.78$\pm$0.32     & 26.02$\pm$13.72     & 20.14$\pm$1.27     & 4.99$\pm$0.07      \\ \hline
$k$NN-UE ($N\textsubscript{probe} = 8$)  & 1.82$\pm$0.28     & 30.18$\pm$16.77     & 20.14$\pm$1.21     & 2.84$\pm$0.08      \\
$k$NN-UE ($N\textsubscript{probe} = 16$) & 1.86$\pm$0.25     & 29.48$\pm$16.91     & 20.13$\pm$1.21     & 3.11$\pm$0.03      \\
$k$NN-UE ($N\textsubscript{probe} = 32$) & 1.92$\pm$0.31     & 28.55$\pm$11.24     & 20.13$\pm$1.22     & 3.31$\pm$0.06      \\
$k$NN-UE ($N\textsubscript{probe} = 64$) & 1.83$\pm$0.28     & 27.00$\pm$9.43      & 20.14$\pm$1.21     & 3.71$\pm$0.06      \\ \hline
& \multicolumn{4}{c}{OntoNotes 5.0 $\mathtt{nw}$ (Out-of-domain)} \\ 
SR                  & 17.05$\pm$0.69    & 37.06$\pm$3.13    & 81.49$\pm$4.17     & 5.75$\pm$0.27      \\
$k$NN-UE (w/o label)  & 8.78$\pm$0.62     & 24.91$\pm$1.81    & 70.10$\pm$4.03     & 10.36$\pm$0.21     \\
$k$NN-UE              & 7.50$\pm$0.42     & 16.53$\pm$2.61    & 74.27$\pm$5.43     & 10.48$\pm$0.12     \\ \hline
$k$NN-UE ($N\textsubscript{probe} = 8$)  & 7.52$\pm$0.41     & 16.01$\pm$1.92    & 74.33$\pm$5.37     & 6.09$\pm$0.28      \\
$k$NN-UE ($N\textsubscript{probe} = 16$) & 7.56$\pm$0.36     & 16.93$\pm$3.38    & 74.31$\pm$5.39     & 6.65$\pm$0.17      \\
$k$NN-UE ($N\textsubscript{probe} = 32$) & 7.60$\pm$0.41     & 17.12$\pm$2.35    & 74.34$\pm$5.35     & 7.33$\pm$0.21      \\
$k$NN-UE ($N\textsubscript{probe} = 64$) & 7.53$\pm$0.40     & 17.28$\pm$2.45    & 74.33$\pm$5.37     & 7.89$\pm$0.12      \\ \hline
\end{tabular}
}
\caption{ECE, MCE, E-AURC and inference time results about NER  on OntoNotes 5.0 $\mathtt{bn}$ (In-domain) and OntoNotes 5.0 $\mathtt{nw}$ (Out-of-domain) for $\mathrm{mDeBERTaV3_{BASE}}$ model when applied IVF in different $N\textsubscript{probe}$.}
\label{tab:clustering_scores}
\end{table}

\subsection{Dimension Reduction}
\label{sec:appendix_analysis_dimension_reduction}
In general, Transformer-based models such as PLM have high-dimensional token representations. In high-dimensional spaces, nearest neighbor search often suffer from the curse of dimensionality. To reduce this problem, we apply dimension reduction to $k$NN-UE similar to \citet{he-etal-2021-efficient}. In this study, we use Principal Component Analysis (PCA) as a dimension reduction algorithm to reduce the dimension of the datastore representations and the query representation $D\textsubscript{pca}$ (In Table~\ref{tab:combination_scores} and ~\ref{tab:combination_coverages}, PCA was performed with $D\textsubscript{pca} = 128$).
As shown in Table~\ref{tab:dimension_reduction_scores}, the UE performance depends on the number of target dimensions, and the performance degrades when $D\textsubscript{pca} = 64$ or $D\textsubscript{pca} = 128$. On the other hand, the performance in $D\textsubscript{pca} = 256$ is almost the same as default $k$NN-UE.
This suggests that excessive dimension reduction in distance computation to extract nearest examples by $k$NN search could have a negative impact on the UE performance.

\begin{table}[t]
\centering
\scalebox{0.56}{
\begin{tabular}{l|cccc}
\hline
\multicolumn{1}{l|}{Methods} & ECE (↓) & MCE (↓) & E-AURC (↓) & time [s] \\ \hline
& \multicolumn{4}{c}{OntoNotes 5.0 $\mathtt{bn}$ (In-domain)} \\ 
SR                  & 7.79$\pm$0.53     & 50.07$\pm$24.15   & 21.90$\pm$1.31     & 2.49$\pm$0.08      \\
$k$NN-UE (w/o label)  & 3.37$\pm$0.71     & 33.15$\pm$3.65    & 17.63$\pm$0.66     & 4.94$\pm$0.10      \\
$k$NN-UE & 1.78$\pm$0.32     & 26.02$\pm$13.72   & 20.14$\pm$1.27     & 4.99$\pm$0.07      \\ \hline
$k$NN-UE ($D\textsubscript{pca} = 64$) & 1.89$\pm$0.37  & 31.01$\pm$14.35  & 20.06$\pm$1.25  & 3.24$\pm$0.08  \\
$k$NN-UE ($D\textsubscript{pca} = 128$) & 1.80$\pm$0.36	  & 27.85$\pm$13.80  & 20.13$\pm$1.29  & 3.41$\pm$0.10  \\
$k$NN-UE ($D\textsubscript{pca} = 256$) & 1.80$\pm$0.40  & 26.23$\pm$12.61  & 20.13$\pm$1.28  & 3.85$\pm$0.06  \\ \hline
& \multicolumn{4}{c}{OntoNotes 5.0 $\mathtt{nw}$ (Out-of-domain)} \\ 
SR                  & 17.05$\pm$0.69    & 37.06$\pm$3.13    & 81.49$\pm$4.17     & 5.75$\pm$0.27      \\
$k$NN-UE (w/o label)  & 8.78$\pm$0.62     & 24.91$\pm$1.81    & 70.10$\pm$4.03     & 10.36$\pm$0.21     \\
$k$NN-UE & 7.50$\pm$0.42     & 16.53$\pm$2.61    & 74.27$\pm$5.43     & 10.48$\pm$0.12     \\ \hline

$k$NN-UE ($D\textsubscript{pca} = 64$) & 7.48$\pm$0.41	  & 16.20$\pm$2.75  & 74.33$\pm$5.49  & 7.37$\pm$0.26  \\
$k$NN-UE ($D\textsubscript{pca} = 128$) & 7.54$\pm$0.45	  & 16.42$\pm$2.73  & 74.30$\pm$5.44  & 7.75$\pm$0.24  \\
$k$NN-UE ($D\textsubscript{pca} = 256$) & 7.56$\pm$0.43	  & 16.13$\pm$2.59  & 74.26$\pm$5.40  & 8.51$\pm$0.46  \\ \hline
\end{tabular}
}
\caption{ECE, MCE, E-AURC and inference time results about NER on OntoNotes 5.0 $\mathtt{bn}$ (In-domain) and OntoNotes 5.0 $\mathtt{nw}$ (Out-of-domain) for $\mathrm{mDeBERTaV3_{BASE}}$ model when applied PCA in different $D\textsubscript{pca}$.}
\label{tab:dimension_reduction_scores}
\end{table}

\section{Distance Recomputation for \texorpdfstring{$k$}{k}NN-UE}
\label{sec:appendix_distance_recomputaton}
When using efficient $k$NN search techniques in Section~\ref{sec:impact_efficient_ann}, we use approximate distances to compute Eq.~\ref{eq:knn_ue_main}. Although we can get raw vectors by using the example indices obtained from approximate nearest neighbor search and compute accurate distance, in $k$NN-LM this has been shown to lead to performance gains and latency degradation~\cite{he-etal-2021-efficient}. We measure the UE performance and inference speed when PQ, clustering, and dimension reduction are applied simultaneously and re-computing accurate distances, reported in Table~\ref{tab:distance_recomputation_score}. These results show that the UE performance does not improve except for MCE in the ID setting, and the latency is about 5-7x slower when reading raw vectors from the datastore and re-computing distances. Moreover, these results suggest that exact distance computation for examples that are not actually nearest neighbors are not very effective in $k$NN-UE.

\begin{table}[t]
\centering
\scalebox{0.56}{
\begin{tabular}{l|cccc}
\hline
\multicolumn{1}{l|}{Methods} & ECE (↓) & MCE (↓) & E-AURC (↓) & time [s] \\ \hline
& \multicolumn{4}{c}{OntoNotes 5.0 $\mathtt{bn}$ (In-domain)} \\ 
$k$NN-UE & 1.78$\pm$0.32     & 26.02$\pm$13.72   & 20.14$\pm$1.27     & 4.99$\pm$0.07      \\ \hline
$k$NN-UE (Approx.) &  2.14$\pm$0.37  & 33.52$\pm$10.84  & 20.12$\pm$1.26	  & 2.87$\pm$0.04  \\
$k$NN-UE (Recomp.) & 2.35$\pm$0.44 & 30.47$\pm$7.50 & 20.16$\pm$1.17 & 16.24$\pm$0.77  \\ \hline
& \multicolumn{4}{c}{OntoNotes 5.0 $\mathtt{nw}$ (Out-of-domain)} \\ 
$k$NN-UE & 7.50$\pm$0.42     & 16.53$\pm$2.61    & 74.27$\pm$5.43     & 10.48$\pm$0.12     \\ \hline

$k$NN-UE (Approx.) & 8.08$\pm$0.53  & 24.03$\pm$5.46  & 74.50$\pm$5.42  & 6.20$\pm$0.20  \\
$k$NN-UE (Recomp.) & 8.30$\pm$0.51 & 25.67$\pm$5.26 & 74.58$\pm$5.53 & 34.22$\pm$0.78 \\ \hline
\end{tabular}
}
\caption{ECE, MCE, E-AURC and inference time results about NER on OntoNotes 5.0 $\mathtt{bn}$ (In-domain) and OntoNotes 5.0 $\mathtt{nw}$ (Out-of-domain) when applying distance recomputation in $k$NN-UE. "Approx." indicates using approximate distances, and "Recomp." indicates using exact distances by distance recomputation. Both "Approx." and "Recomp." are applied PQ with $N\textsubscript{sub} = 32$, clustering with $N\textsubscript{probe} = 32$ and dimension reduction with $D\textsubscript{pca} = 128$.}
\label{tab:distance_recomputation_score}
\end{table}

\section{Licenses of Datasets, Tools and Models}
\label{sec:appendix_licenses}
\paragraph{Datasets} The IMDb movie dataset can be used for research purposes as described in \url{https://developer.imdb.com/non-commercial-datasets/}. Yelp dataset can be used for academic purposes as described in~\url{https://s3-media0.fl.yelpcdn.com/assets/srv0/engineering_pages/f64cb2d3efcc/assets/vendor/Dataset_User_Agreement.pdf}. The MNLI dataset is licensed for research purposes as described in~\citet{williams-etal-2018-broad}. The SNLI dataset can be used for research purposes as described in~\url{https://nlp.stanford.edu/projects/snli/}. OntoNotes 5.0 dataset can be used for research purposes as described in \url{https://catalog.ldc.upenn.edu/LDC2013T19}.

\paragraph{Tools} $\mathtt{transformers}$ is licensed by Apache-2.0. $\mathtt{faiss}$ is MIT-licensed.

\paragraph{Models} $\mathrm{DeBERTaV3_{BASE}}$ and $\mathrm{mDeBERTaV3_{BASE}}$ from Huggingface model checkpoints are MIT-licensed.

\end{document}